\documentclass{article}
\usepackage{graphicx} 
\usepackage{aaai25}  
\usepackage{times}  
\usepackage{helvet}  
\usepackage{courier}  
\usepackage[hyphens]{url}  
\usepackage{natbib}  
\usepackage{caption} 
\usepackage{algorithm}
\usepackage{algorithmic}
\usepackage{xcolor}
\usepackage{fancyhdr}
\usepackage{newfloat}
\usepackage{listings}
\DeclareCaptionStyle{ruled}{labelfont=normalfont,labelsep=colon,strut=off} 
\floatstyle{ruled}
\newfloat{listing}{tb}{lst}{}
\floatname{listing}{Listing}
\fancypagestyle{firstpage}{
  \fancyhf{}  
  \fancyfoot[C]{1st Workshop on Preparing Good Data for Generative AI: Challenges and Approaches@ AAAI 2025}  
}

\title{FinGPT: Enhancing Sentiment-Based Stock Movement Prediction with Dissemination-Aware and Context-Enriched LLMs
}
\author{
    Yixuan Liang$^{1,2}$, Yuncong Liu$^{1,2}$, Neng Wang$^{1}$, Hongyang Yang$^{1,2}$, \\ Boyu Zhang$^{1*}$, Christina Dan Wang$^{1,3}$\thanks{Corresponding author}
}

\affiliations{
    $^1$AI4Finance Foundation \\
    $^2$Columbia University\\
    $^3$New York University Shanghai \\
    contact@ai4finance.org
}

\begin{document}

\maketitle

\begin{abstract}

Financial sentiment analysis is crucial for understanding the influence of news on stock prices. Recently, large language models (LLMs) have been widely adopted for this purpose due to their advanced text analysis capabilities. However, these models often only consider the news content itself, ignoring its dissemination, which hampers accurate prediction of short-term stock movements. Additionally, current methods often lack sufficient contextual data and explicit instructions in their prompts, limiting LLMs' ability to interpret news. In this paper, we propose a data-driven approach that enhances LLM-powered sentiment-based stock movement predictions by incorporating news dissemination breadth, contextual data, and explicit instructions. We cluster recent company-related news to assess its reach and influence, enriching prompts with more specific data and precise instructions. This data is used to construct an instruction tuning dataset to fine-tune an LLM for predicting short-term stock price movements. Our experimental results show that our approach improves prediction accuracy by 8\% compared to existing methods.

\end{abstract}

\section{Introduction}
Financial markets are highly reactive to news, social media, and other public sentiment signals; these affect trading behaviors and, ultimately, stock prices. Understanding these sentiment shifts can provide valuable insights into price movement patterns, making sentiment analysis an essential component of modern financial forecasting.

Traditional sentiment analysis typically categorizes sentiment as positive, negative, or neutral. Advances in natural language processing (NLP) have significantly enhanced our ability to analyze and interpret sentiment data from vast text sources. Most prior research has focused on improving the accuracy of sentiment analysis for individual news items, rarely aggregating them to assess the overall market sentiment or integrating the results into downstream tasks such as stock prediction and risk management.

The emergence of Large Language Models (LLMs) has revolutionized financial sentiment analysis by providing not only sentiment-based classification but also explanations for stock movement predictions \cite{zhang2023sentiment,zhang2023enhancing,araci2019finbert,wu2023bloomberggpt}. Recent works like FinRobot \cite{yang2024finrobot,zhou2024finrobot,han2024enhancing} demonstrate this capability through their ``Market Forecaster'' tool, which moves beyond single-news analysis to capture more breadth sentiment landscapes. Additionally, we are witnessing a growing body of research that extends beyond individual news analysis \cite{wang2024llmfactor}, aiming to offer a more holistic view of stock market dynamics.


Despite the advancements brought by LLMs in financial sentiment analysis, existing methods often rely solely on the news content itself for predictions. This approach neglects the crucial factor of news dissemination, which significantly affects market reactions and stock price movements. Additionally, these methods often lack sufficient contextual data and explicit instructions, limiting the LLMs' ability to interpret news. Our proposed approach addresses these limitations by incorporating the breadth of news dissemination, detailed contextual data, and precise instructions, thereby enhancing the accuracy of short-term stock price movement predictions.



In this paper, we propose a novel approach to enhance LLM-powered sentiment-based stock movement predictions by incorporating news dissemination breadth, contextual data, and precise instructions. Our methodology clusters recent company-related news articles, using cluster attributes to evaluate the news's reach and influence. We operate under two key assumptions: \textit{i)} The centroid article of each cluster encapsulates the most comprehensive information for LLM processing; \textit{ii)} The cluster size indicates the topic's market impact, with larger clusters signifying more significant events. Additionally, we enhance the prompts with daily stock price and return data, along with instructions to consider the short-term or long-term impact of the news. Utilizing this information, we construct an instruction tuning dataset to fine-tune an LLM for short-term stock price predictions.



We summarize by our key contributions:
\begin{enumerate}
    \item We propose a data-driven clustering-based method to capture the breadth of news dissemination, and incorporate it into the training dataset.

    \item By enriching prompts with contextual data and instructions tailored to our proposed data format, we offer a more nuanced approach to financial sentiment analysis.

    \item Our experimental results demonstrate that our approach improves prediction accuracy by 8\% compared to existing methods, offering a more robust and efficient framework for understanding the impact of news on stock prices.
    


\end{enumerate}

The remainder of the paper is structured as follows: Section 2 reviews related work in sentiment-based financial prediction. Section 3 presents the problem statement. Section 4 describes our data-centric methodology, including high-granularity and news clustering methods. Section 5 discusses performance metrics, evaluation results, and a case study on Boeing. Section 6 concludes with findings and future directions.

\section{Related Works}
\subsection{NLP for Financial Sentiment Analysis}
Sentiment analysis has long been a key application of natural language processing (NLP), especially in finance, where it provides valuable insights into market trends and investor sentiment \cite{chan2017sentiment, atkins2018financial}. Various models and methodologies have been developed to enhance the accuracy and efficiency of sentiment analysis \cite{tai2013automatic, hamilton2016inducing, day2016deep, sohangir2018big, mishev2020evaluation, rizinski2024sentiment}, ranging from lexicon-based techniques to machine learning and deep learning approaches.

Recent advances in large language models (LLMs) have demonstrated remarkable capabilities in understanding complex natural language, with promising applications in financial sentiment analysis (e.g. FinBERT \cite{araci2019finbert}, Bloomberggpt \cite{wu2023bloomberggpt}, Fingpt \cite{wang2023fingpt,yang2023fingpt}).





While these financial language models excel at individual news sentiment analysis, the systematic integration of related news for stock price prediction remains largely unexplored.

\subsection{LLMs for Sentiment-based Stock Price Prediction}
Leveraging text or news for stock prediction is not new; prior work has used tweets and historical data to forecast prices \cite{xu2018stock}. Now, with the advent of LLMs, we can achieve more nuanced understanding and interpretation of financial text, allowing these models to capture complex relationships within news data and better inform stock price predictions.

Recent works demonstrate diverse applications of LLMs in stock prediction. LLMFactor \cite{wang2024llmfactor} targets short-term prediction through Sequential Knowledge-Guided Prompting, providing real-time interpretable insights. Similarly, \cite{elahi2024combining} addresses longer-term predictions by combining financial data and news through retrieval-augmented techniques for 3-6 month horizons.

Our work builds on this line of work, but are independent of the model itself: we focus exclusively on the data preparation process, incorporating the impact of news dissemination on stock price movements and providing LLMs with more precise instructions. In this paper, we follow the standard framework of instruction tuning LLMs for financial forecasting and used the data organization as the baseline (outlined in \cite{yang2024finrobot}) to show the significant improvement (Section 4), but we believe that our methodology has a much broader application, having the potential to be applied to all the existing models.

\section{Problem Setting and Overall Framework}

Our objective is to predict weekly stock price movements based on news sentiment. The movements are categorized into twelve labels: U1-U5 and U5+ for upward trends (0-1\%, 1-2\%, 2-3\%, 3-4\%, 4-5\%, over 5\%), and D1-D5+ for corresponding downward trends. Predictions are based on previous week's stock prices, recent news, and company fundamentals (updated quarterly and included three weeks after the quarterly report release). The model also generates reasoning for the prediction by identifying [Positive Developments] and [Potential Concerns]—highlighting the 2-4 most significant factors in each category—as well as providing [Prediction \& Analysis].

\begin{figure}[H]
    \centering
    \includegraphics[width=0.41\textwidth]{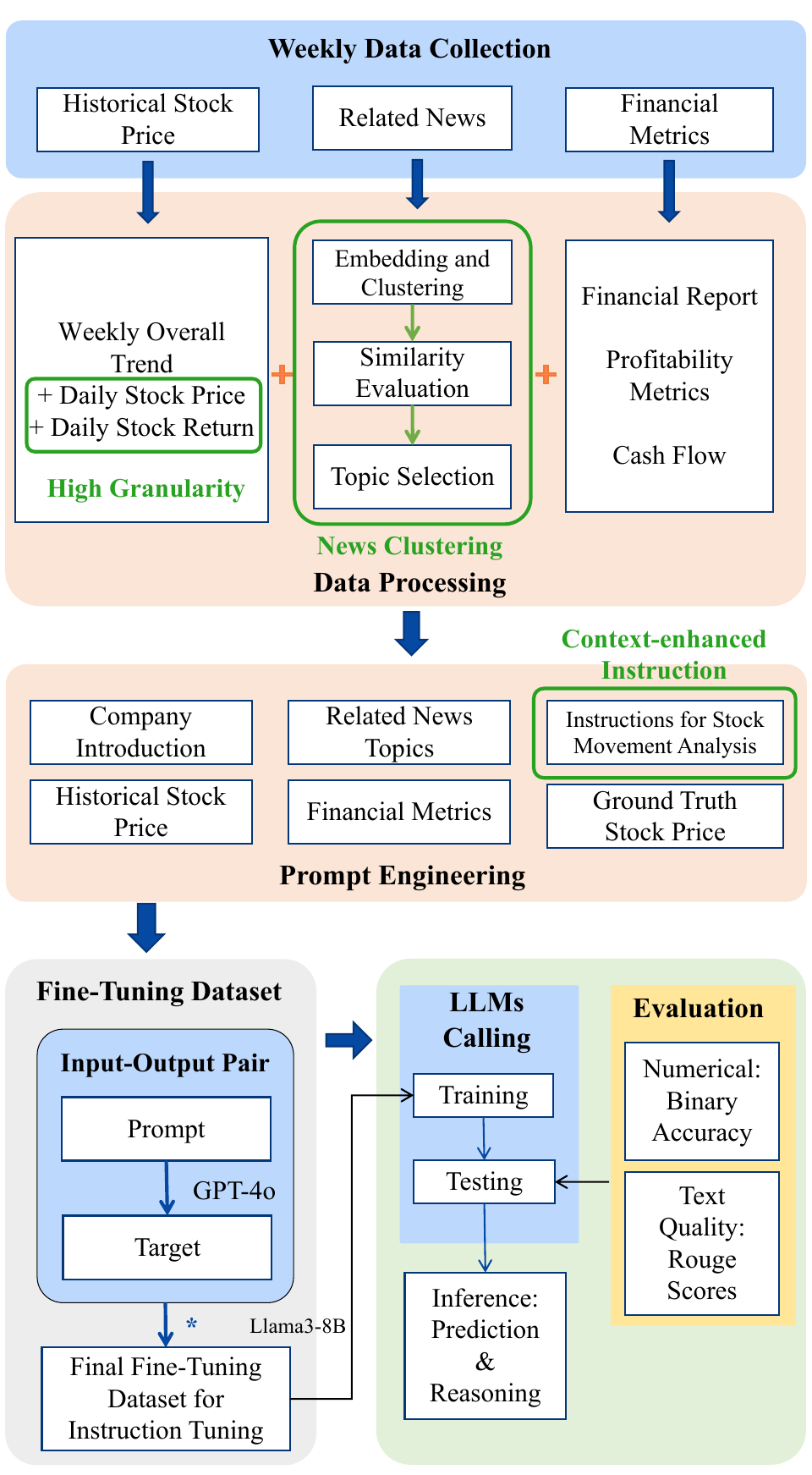}
    \caption{Overall framework. In Data Processing part, we increase granularity in the historical stock price data and employ news data clustering. In Prompt Engineering part, we incorporate context-enhanced instructions for stock movement analysis. The operation we mark with ``*'' is removing future movement label.}
    \label{fig1}
\end{figure}

Our overall framework is illustrated in Figure \ref{fig1}, following a standard framework of fine-tuning LLMs for financial analysis. Our work focuses on the Data Processing part and Prompt Engineering part in this flow.

\section{Methodology}
This section outlines our data-driven methodologies. Specifically, we 1) increase stock price granularity and implemented news clustering in the Data Processing part, and 2) incorporate contextual and more targeted instructions in Prompt Engineering part. 

\subsection{Data Processing}
\subsubsection{High Granularity in Stock Price Information (HG): }The baseline method uses only weekly aggregate stock price movements (e.g. 3\% weekly gain). To enhance prediction performance, we increase data granularity by incorporating daily closing prices and corresponding returns throughout each week. This granular approach serves two key purposes: 1) it reduces the uncertainty of the calculation within the LLMs by providing explicit daily price movements and 2) enables precise temporal alignment between price changes and news events, providing a basis for the differentiation of short- long-term impacts. We will refer to this method as \textbf{HG} in the following part.

\subsubsection{News Clustering (HG-NC): }
Traditional stock market analysis often lacks systematic quantification of news dissemination. A more comprehensive approach requires analyzing the complete news landscape-often exceeding 200 articles weekly for active stocks. It presents significant challenges: redundant information processing, computational inefficiency, and potential token limitations in language models. 

To address these challenges, built upon our HG method, we further developed a clustering approach that efficiently organizes high-volume news content while capturing news impact through two key dimensions: reporting frequency and temporal span. We will refer to this method as \textbf{HG-NC} in the following part. Our approach comprises the following steps:

\begin{enumerate}
    \item \textbf{Data Collection}: We retrieve weekly financial news data, including titles and summaries, from the Finnhub API.

    \item \textbf{Topic Clustering}: News articles are transformed into embedding representations using Sentence Transformers, followed by BERTopic-based topic modeling to identify and cluster thematically related news content.

    \item \textbf{Cluster Quality Assessment}: We evaluate cluster cohesiveness through pairwise similarity analysis:
        \begin{itemize}
        \item \textbf{High-Cohesion Clusters} (average pairwise similarity $>$ 0.6): For these clusters, we select the centroid-proximate article as the cluster representative and preserve the metadata including cluster size and temporal span.
    
        \item \textbf{Low-Cohesion Clusters} (average pairwise similarity $\leq$ 0.6): we again select the article closest to the centroid but limit the topic size to 2 and record the time range. This setting reflects lower clustering quality and lower confidence in these less cohesive groups.
    \end{itemize}

    \item \textbf{Topic Selection Strategy}: When high-cohesion clusters fall below six, we supplement with low-cohesion clusters (at most 4) to ensure sufficient information coverage. All parameters—similarity thresholds, cluster sizes, and topic quotas—are adjustable based on analysis needs and LLM constraints.

\end{enumerate}

The clustering approach, leveraging BERTopic \cite{grootendorst2022bertopic} and cosine similarity evaluation, efficiently condenses large volumes of news into representative samples while quantifying news dissemination, thereby enhancing stock movement prediction. 

\subsection{Prompt Engineering: Context-enhanced Instructions}

To adapt to our proposed data format where we incorporate daily stock information and quantified news dissemination, context-enhanced instructions are needed for better analysis. 

\subsubsection {For HG:} We instruct LLMs to differentiate between short-term and long-term impacts of news, as daily stock prices and returns can reveal immediate market reactions. This distinction is crucial because the influence of short-term news is often already reflected in stock movements within the same week. For a prompt template, see Figure \ref{template2}.

\subsubsection{For HG-NC:} Built upon HG, we construct the news component using selected representative articles and their associated metadata (topic size and temporal coverage). Then, we enhance the instructions by describing the news component and providing guidelines for analyzing the impact of news dissemination on stock movement. For a prompt template, see Figure \ref{template3}.

\subsection{Instruction-tuning}
The training dataset pairs our structured input prompts (including company introduction, historical stock prices, related news, company fundamentals, and instructions for utilizing sentiment analysis for prediction) with GPT-4o-generated analysis based on known future movements. Then, removing the ground truth stock price in the prompt, we use this dataset to fine-tune Llama3-8B for weekly stock movement prediction and evaluate both numerical accuracy and reasoning quality (see Appendix A for details.)

\section{Performance Evaluation}

We evaluate our models' performance using two key metrics: binary accuracy for stock prediction and ROUGE scores for reasoning quality. Our analysis compares three progressive methods: the \textbf{baseline}, the \textbf{HG} method with increased stock price granularity, and the \textbf{HG-NC} method.

\subsection{Binary Accuracy in Stock Movement Prediction}
We assess each model's ability to predict directional stock price movements (up/down) using binary accuracy metrics. Our comprehensive dataset comprises 380 observations across 20 companies, spanning multiple market sectors to ensure robust evaluation. The results demonstrate a consistent improvement pattern across our model iterations:

\begin{table}[htb]
\centering
\small
\renewcommand{\arraystretch}{1.2}
\begin{tabular}{|c|c|c|c|c|}
\hline
\textbf{Method}        & \textbf{Avg Acc} & \textbf{Long term} & \textbf{Short term} \\ \hline
Baseline        & 55.0\%            & 15.0\%                 & 7.5\%                  \\ \hline
HG       & 59.2\%            & 69.8\%                 & 56.6\%                 \\ \hline
HG-NC        & 63.0\%            & 58.5\%                    & 50.9\%                   \\ \hline
\end{tabular}
\caption{Our methods show a significant increase in the binary accuracy. Also, the increase in ``long-term" word frequency and ``short-term"  word frequency provides evidence for attention to temporal aspects, which may account for the accuracy improvement.
}
\label{tab:methods_accuracy}
\end{table}


With high granularity stock price and targeted instructions, accuracy improves from 0.550 to 0.592. A detailed analysis reveals increased attention to temporal aspects in the \textbf{Prediction \& Analysis} component, with the frequency of ``long-term" rising from 15.0\% to 69.8\% and ``short-term" from 7.5\% to 56.6\%. This indicates that the LLM effectively follows the instructions to differentiate between short-term and long-term news impacts, balancing them in stock price predictions, leading to improved accuracy.

The further increase to 63\% with NC validates our hypothesis that incorporating news clustering results enhances the LLM's capacity to capture market dynamics and the impact of news dissemination on stock movements.

\subsection{ROUGE Scores for Reasoning Quality Evaluation}
Given the large size of our training and test datasets, obtaining ground-truth sentiment-based analysis for every instance is impractical. Therefore, we rely on automated evaluation metrics to assess the quality of model outputs. We use ROUGE (Recall-Oriented Understudy for Gisting Evaluation) scores \cite{lin2004rouge} to assess the reasoning quality as they measure the overlap of key words and phrases between LLM-generated outputs and reference summaries. Higher ROUGE scores indicate closer alignment with the reference text and broader coverage of news content as the basis for reasoning. However, precise and comprehensive evaluation still requires human judgment.


We evaluate three metrics: ROUGE-1, ROUGE-2, and ROUGE-L (longest common subsequences), with ROUGE-N representing N-gram co-occurrence statistics. Across all metrics, the HG-NC method consistently outperforms both the baseline and HG approaches.
\begin{table}[h]
\centering
\setlength{\tabcolsep}{4pt}
\renewcommand{\arraystretch}{1.2}
\small
\begin{tabular}{|c|c|c|c|}
\hline
\textbf{Method}          & \textbf{ROUGE-1} & \textbf{ROUGE-2} & \textbf{ROUGE-L} \\ \hline
Baseline & 0.450          & 0.121         & 0.224         \\ \hline
HG & 0.469          & 0.131          & 0.224          \\ \hline
HG-NC & 0.472         & 0.140          & 0.234          \\ \hline
\end{tabular}
\caption{ROUGE scores for different methods}
\label{tab:rouge_results}
\end{table}

Specifically, we analyze ROUGE scores for the \textbf{Prediction \& Analysis} component in the output, which integrates positive and negative factors to justify the model's directional predictions. This critical component highlights the LLM's ability to weigh competing factors from sentiment analysis and articulate its decision-making process. As shown in Table~\ref{tab:rouge_results}, the HG-NC method better captures and articulates the complex interactions between various market factors in its analysis.

\subsection{Case Study: Boeing Company}
We use The Boeing Company (NYSE: BA) as a representative case study and compare the prediction performance of the HG method and HG-NC method. Overall, the HG-NC method has accuracy (63.2\%) compared to the HG method (52.63\%). For prediction results see Figure \ref{label} in Appendix.

\begin{figure}[H]
    \centering
    \includegraphics[width=0.44\textwidth]{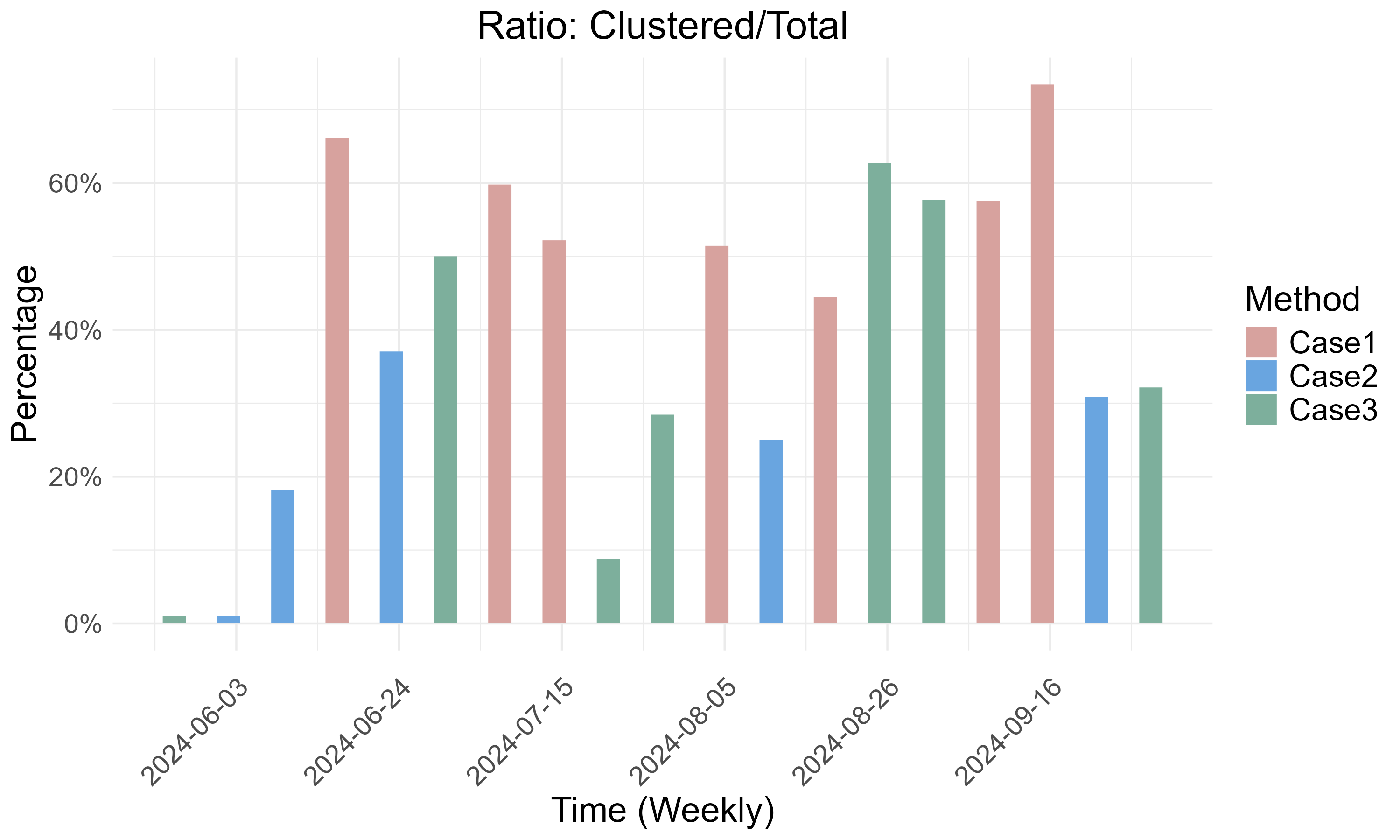}
    \caption{The ratio of news articles in high-coherence clusters to the total number of news articles. We classify prediction comparisons into three cases: Case 1: HG-NC correct vs. NC incorrect; Case 2: HG correct vs. HG-NC incorrect; Case 3: Both methods yield identical predictions.}
    \label{ratio}
\end{figure}

We examine the ratio of news articles in high-coherence clusters (those with average pairwise similarity $>$ 0.6) to the total number of news articles as an indicator of clustering performance. In general, we observe a strong correlation between clustering performance and prediction performance. Specifically, as shown in Figure \ref{ratio}, in 7 instances where our HG-NC method outperforms HG (Case 1), we observe relatively high ratios of high-coherence clusters (mostly exceeding 50\%). Conversely, performance declines when this ratio falls below 40\% (Case 2), suggesting insufficient capture or preservation of significant market information.

\section{Conclusion}

In this paper, we proposed an approach to enhance LLM-powered sentiment-based stock movement prediction. By increasing granularity for stock price and providing instructions for short-term or long-term analysis, we enhance the contextual understanding of news. Further, we evaluate news dissemination through clustering and incorporate its market impact to improve predictions. We developed an instruction tuning dataset to fine-tune LLMs for more accurate short-term stock movement predictions. Experimental results validate our approach, achieving 63\% binary accuracy compared to the 55\% baseline, with better predictions at high clustering ratios ($>50\%$). These findings highlight the importance of enriched contextual data and dissemination-aware methods in improving prediction accuracy.


\bibliography{bib}

\begin{thebibliography}{22}
\providecommand{\natexlab}[1]{#1}

\bibitem[{Araci(2019)}]{araci2019finbert}
Araci, D. 2019.
\newblock FinBERT: Financial Sentiment Analysis with Pre-trained Language Models.
\newblock \emph{arXiv preprint arXiv:1908.10063}.

\bibitem[{Atkins, Niranjan, and Gerding(2018)}]{atkins2018financial}
Atkins, A.; Niranjan, M.; and Gerding, E. 2018.
\newblock Financial news predicts stock market volatility better than close price.
\newblock \emph{The Journal of Finance and Data Science}, 4(2): 120--137.

\bibitem[{Chan and Chong(2017)}]{chan2017sentiment}
Chan, S.~W.; and Chong, M.~W. 2017.
\newblock Sentiment analysis in financial texts.
\newblock \emph{Decision Support Systems}, 94: 53--64.

\bibitem[{Day and Lee(2016)}]{day2016deep}
Day, M.-Y.; and Lee, C.-C. 2016.
\newblock Deep learning for financial sentiment analysis on finance news providers.
\newblock In \emph{2016 IEEE/ACM International Conference on Advances in Social Networks Analysis and Mining (ASONAM)}, 1127--1134. IEEE.

\bibitem[{Elahi and Taghvaei(2024)}]{elahi2024combining}
Elahi, A.; and Taghvaei, F. 2024.
\newblock elahi2024combiningCombining Financial Data and News Articles for Stock Price Movement Prediction Using Large Language Models.
\newblock \emph{arXiv preprint arXiv:2411.01368}.

\bibitem[{Grootendorst(2022)}]{grootendorst2022bertopic}
Grootendorst, M. 2022.
\newblock BERTopic: Neural topic modeling with a class-based TF-IDF procedure.
\newblock \emph{arXiv preprint arXiv:2203.05794}.

\bibitem[{Hamilton et~al.(2016)Hamilton, Clark, Leskovec, and Jurafsky}]{hamilton2016inducing}
Hamilton, W.~L.; Clark, K.; Leskovec, J.; and Jurafsky, D. 2016.
\newblock Inducing domain-specific sentiment lexicons from unlabeled corpora.
\newblock In \emph{Proceedings of the conference on empirical methods in natural language processing. conference on empirical methods in natural language processing}, volume 2016, 595. NIH Public Access.

\bibitem[{Han et~al.(2024)Han, Wang, Che, Yang, Zhang, and Xu}]{han2024enhancing}
Han, X.; Wang, N.; Che, S.; Yang, H.; Zhang, K.; and Xu, S.~X. 2024.
\newblock Enhancing Investment Analysis: Optimizing AI-Agent Collaboration in Financial Research.
\newblock In \emph{ICAIF 2024: Proceedings of the 5th ACM International Conference on AI in Finance}, 538--546.

\bibitem[{Lin(2004)}]{lin2004rouge}
Lin, C.-Y. 2004.
\newblock Rouge: A package for automatic evaluation of summaries.
\newblock In \emph{Text summarization branches out}, 74--81.

\bibitem[{Mishev et~al.(2020)Mishev, Gjorgjevikj, Vodenska, Chitkushev, and Trajanov}]{mishev2020evaluation}
Mishev, K.; Gjorgjevikj, A.; Vodenska, I.; Chitkushev, L.~T.; and Trajanov, D. 2020.
\newblock Evaluation of sentiment analysis in finance: from lexicons to transformers.
\newblock \emph{IEEE access}, 8: 131662--131682.

\bibitem[{Rizinski et~al.(2024)Rizinski, Peshov, Mishev, Jovanovik, and Trajanov}]{rizinski2024sentiment}
Rizinski, M.; Peshov, H.; Mishev, K.; Jovanovik, M.; and Trajanov, D. 2024.
\newblock Sentiment Analysis in Finance: From Transformers Back to eXplainable Lexicons (XLex).
\newblock \emph{IEEE Access}.

\bibitem[{Sohangir et~al.(2018)Sohangir, Wang, Pomeranets, and Khoshgoftaar}]{sohangir2018big}
Sohangir, S.; Wang, D.; Pomeranets, A.; and Khoshgoftaar, T.~M. 2018.
\newblock Big Data: Deep Learning for financial sentiment analysis.
\newblock \emph{Journal of Big Data}, 5(1): 1--25.

\bibitem[{Tai and Kao(2013)}]{tai2013automatic}
Tai, Y.-J.; and Kao, H.-Y. 2013.
\newblock Automatic domain-specific sentiment lexicon generation with label propagation.
\newblock In \emph{Proceedings of international conference on information integration and web-based applications \& services}, 53--62.

\bibitem[{Wang, Izumi, and Sakaji(2024)}]{wang2024llmfactor}
Wang, M.; Izumi, K.; and Sakaji, H. 2024.
\newblock LLMFactor: Extracting Profitable Factors through Prompts for Explainable Stock Movement Prediction.
\newblock \emph{arXiv preprint arXiv:2406.10811}.

\bibitem[{Wang, Yang, and Wang(2023)}]{wang2023fingpt}
Wang, N.; Yang, H.; and Wang, C.~D. 2023.
\newblock Fingpt: Instruction tuning benchmark for open-source large language models in financial datasets.
\newblock \emph{arXiv preprint arXiv:2310.04793}.

\bibitem[{Wu et~al.(2023)Wu, Irsoy, Lu, Dabravolski, Dredze, Gehrmann, Kambadur, Rosenberg, and Mann}]{wu2023bloomberggpt}
Wu, S.; Irsoy, O.; Lu, S.; Dabravolski, V.; Dredze, M.; Gehrmann, S.; Kambadur, P.; Rosenberg, D.; and Mann, G. 2023.
\newblock Bloomberggpt: A large language model for finance.
\newblock \emph{arXiv preprint arXiv:2303.17564}.

\bibitem[{Xu and Cohen(2018)}]{xu2018stock}
Xu, Y.; and Cohen, S.~B. 2018.
\newblock Stock movement prediction from tweets and historical prices.
\newblock In \emph{Proceedings of the 56th Annual Meeting of the Association for Computational Linguistics (Volume 1: Long Papers)}, 1970--1979.

\bibitem[{Yang, Liu, and Wang(2023)}]{yang2023fingpt}
Yang, H.; Liu, X.-Y.; and Wang, C.~D. 2023.
\newblock Fingpt: Open-source financial large language models.
\newblock \emph{arXiv preprint arXiv:2306.06031}.

\bibitem[{Yang et~al.(2024)Yang, Zhang, Wang, Guo, Zhang, Lin, Wang, Zhou, Guan, Zhang et~al.}]{yang2024finrobot}
Yang, H.; Zhang, B.; Wang, N.; Guo, C.; Zhang, X.; Lin, L.; Wang, J.; Zhou, T.; Guan, M.; Zhang, R.; et~al. 2024.
\newblock FinRobot: An Open-Source AI Agent Platform for Financial Applications using Large Language Models.
\newblock \emph{arXiv preprint arXiv:2405.14767}.

\bibitem[{Zhang et~al.(2023{\natexlab{a}})Zhang, Yang, Zhou, Ali~Babar, and Liu}]{zhang2023enhancing}
Zhang, B.; Yang, H.; Zhou, T.; Ali~Babar, M.; and Liu, X.-Y. 2023{\natexlab{a}}.
\newblock Enhancing financial sentiment analysis via retrieval augmented large language models.
\newblock In \emph{Proceedings of the fourth ACM international conference on AI in finance}, 349--356.

\bibitem[{Zhang et~al.(2023{\natexlab{b}})Zhang, Deng, Liu, Pan, and Bing}]{zhang2023sentiment}
Zhang, W.; Deng, Y.; Liu, B.; Pan, S.~J.; and Bing, L. 2023{\natexlab{b}}.
\newblock Sentiment analysis in the era of large language models: A reality check.
\newblock \emph{arXiv preprint arXiv:2305.15005}.

\bibitem[{Zhou et~al.(2024)Zhou, Wang, Wu, and Yang}]{zhou2024finrobot}
Zhou, T.; Wang, P.; Wu, Y.; and Yang, H. 2024.
\newblock FinRobot: {AI} Agent for Equity Research and Valuation with Large Language Models.
\newblock In \emph{ICAIF 2024: The 1st Workshop on Large Language Models and Generative AI for Finance}.

\end{thebibliography}

\clearpage

\appendix

\section*{Appendix A: Model Training}
The training parameters are given in Table. For our model, we initialize it with LLAMA-3-8B model and perform instruction tuning over 5 epochs. The training process utilizes the AdamW optimizer, with a batch size of 32, an initial learning rate of $1 \times 10^{-5}$, and a weight decay of 0.01. In order to be able to input the normal amount of news in a week, we set a maximum input text length of 8000 tokens. We utilize DeepSpeed for the fine-tuning process on A100 (40GB) GPU, resulting in a total training time of 162 minutes.

\begin{table}[h!]
\centering
\renewcommand{\arraystretch}{1.2}
\begin{tabular}{|l|c|}
\hline
\textbf{Parameter} & \textbf{Value} \\ \hline
Learning rate      & $1 \times 10^{-5}$           \\ \hline
Weight Decay       & 0.01            \\ \hline
Batch size         & 32             \\ \hline
Training epochs    & 5             \\ \hline
Num warmup Steps   & 0              \\ \hline
Max Token Length   & 8000            \\ \hline
GPUs               & A100 (40GB)  \\ \hline
\end{tabular}
\caption{Training parameters.}
\label{table:training_parameters}
\end{table}

\section{Appendix B: Prompt Templates}

\begin{figure}[htb]
    \centering
    \includegraphics[width=0.45\textwidth]{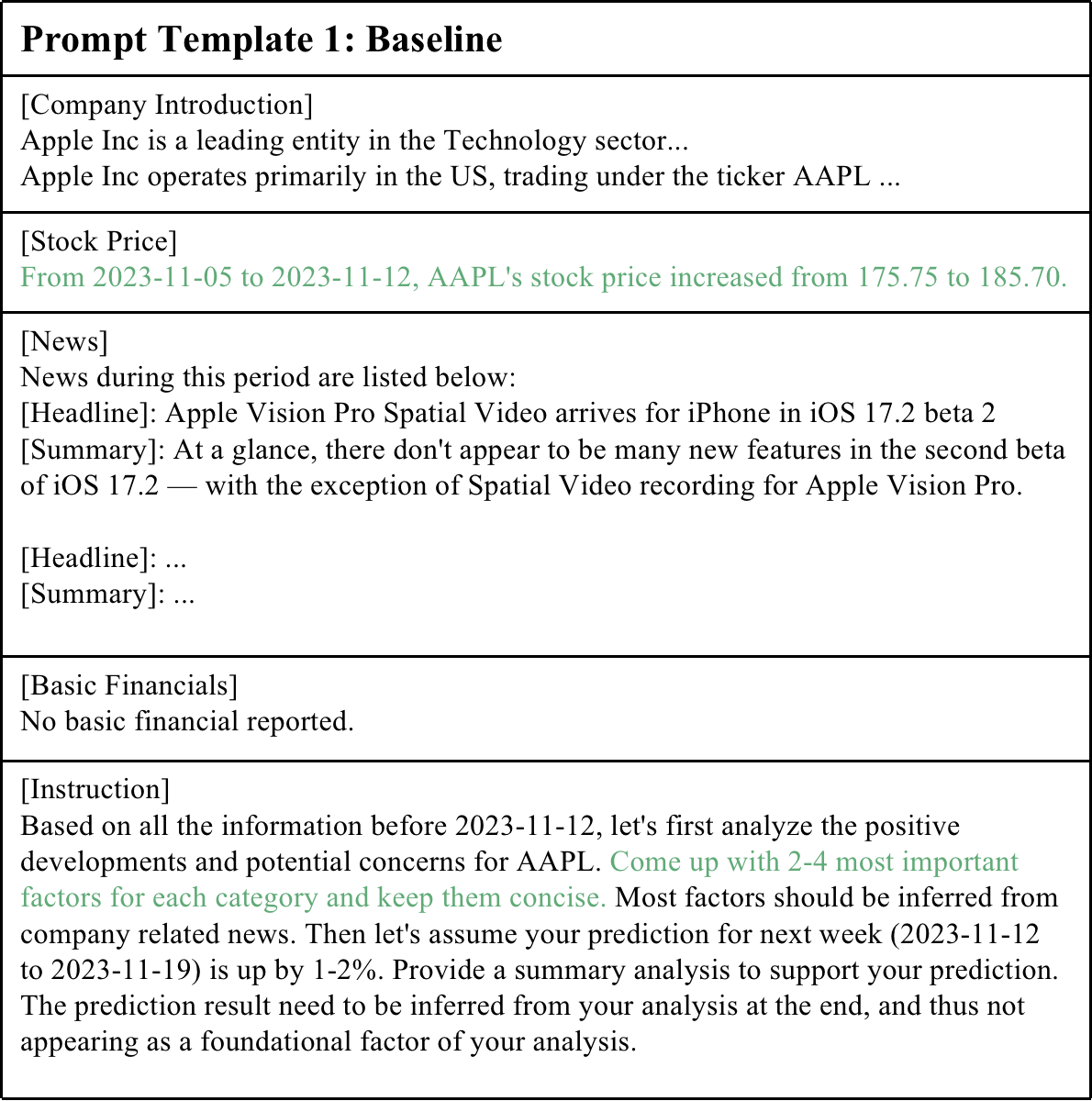}
    \caption{Baseline Prompt Template. This template contains company introduction, stock price weekly trend, news headlines and summaries, basic financials, and analytical instruction for prediction.}
    \label{template1}
\end{figure}

\begin{figure}[!htb]
    \centering
    \includegraphics[width=0.45\textwidth]{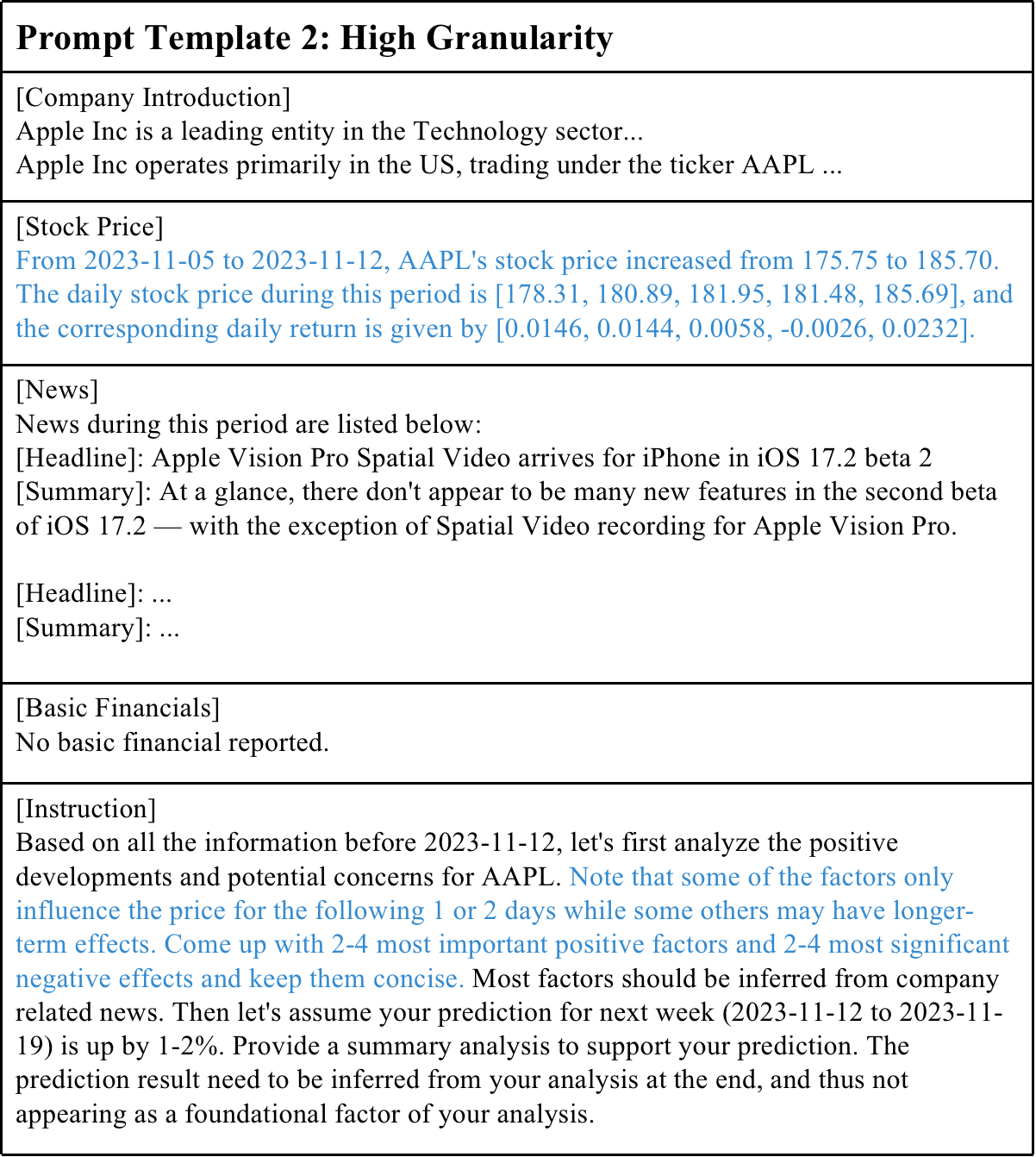}
    \caption{HG Prompt Template. This template includes daily stock prices with corresponding returns and context-enhanced analytical instructions emphasizing temporal effects.}
    \label{template2}
\end{figure}

\begin{figure}[htb]
    \centering
    \includegraphics[width=0.45\textwidth]{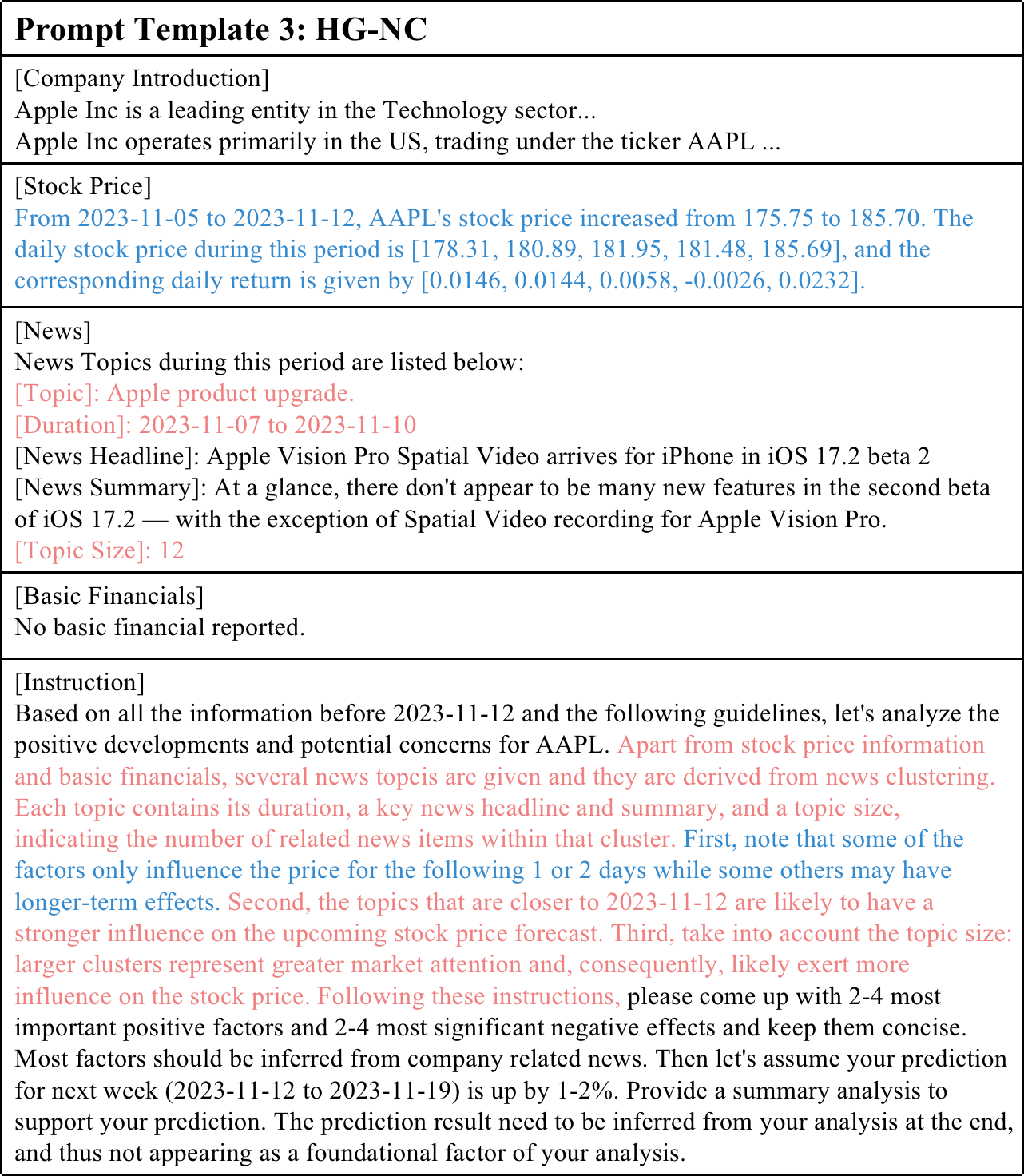}
    \caption{HG-NC Prompt Template. This template enhances the HG format by incorporating clustered news topics with topic size and temporal span, along with guidelines for analyzing news dissemination on price predictions.}
    \label{template3}
\end{figure}

\clearpage

\section{Appendix C: Experimental results for Boeing Company}

\begin{figure}[H]
\vspace{+1.5 cm}
    \centering
    \includegraphics[width=0.47\textwidth]{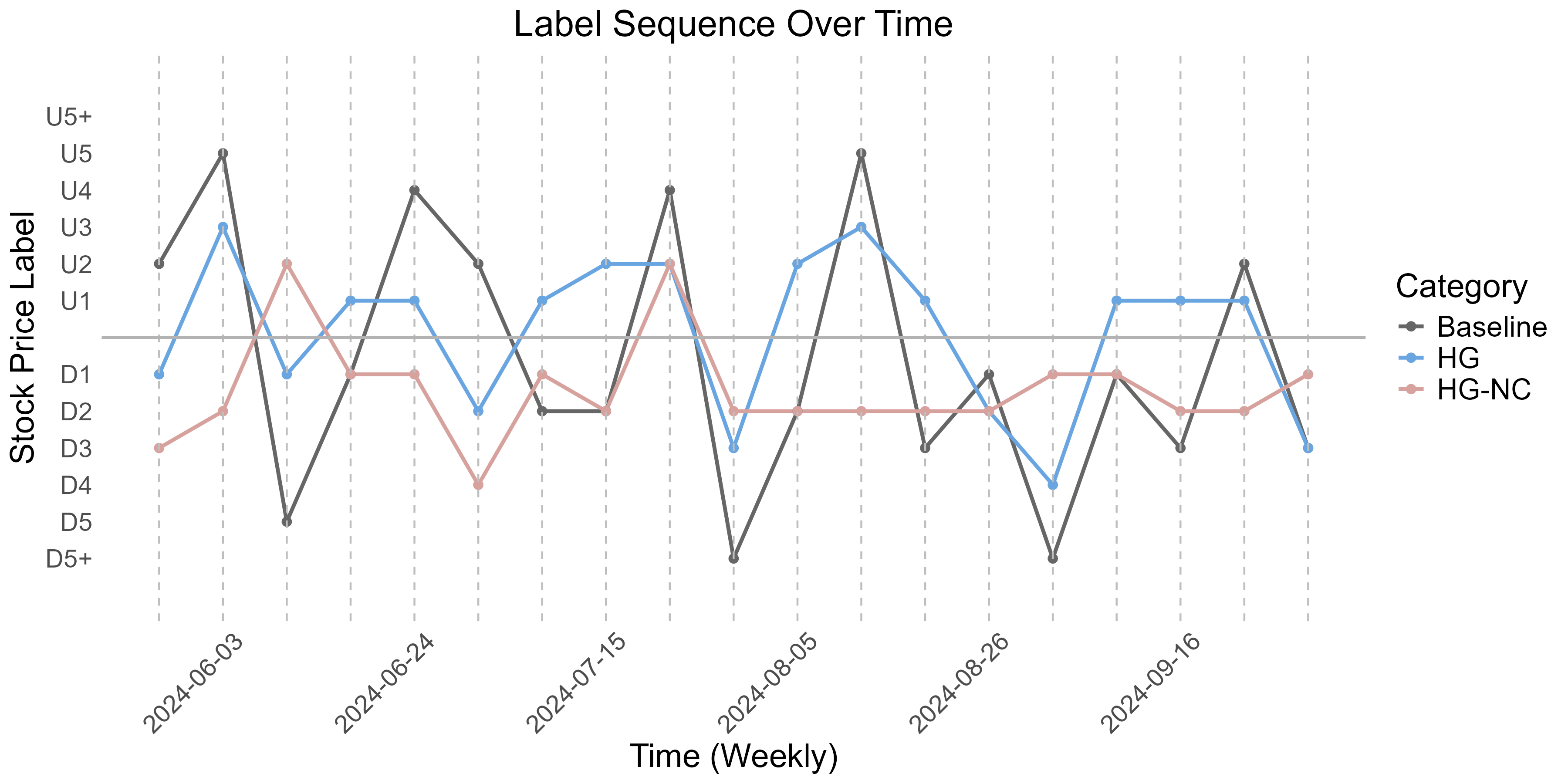}
    \caption{BA prediction labels during test period.}
    \label{label}
\end{figure}

\begin{table}[h]
\vspace{+1 cm}
\small
\begin{tabular}{|c|c|c|c|c|c|}
\hline
Start & News &  & Good & Clustered &  \\
Date & Count & Clusters & Clusters & News & Ration \\
\hline
5/26 & 74 & 2 & 0 & 0 & 0 \\
6/2 & 76 & 4 & 1 & 3 & 0.04 \\
6/9 & 77 & 5 & 1 & 14 & 0.18 \\
6/16 & 118 & 9 & 5 & 78 & 0.66 \\
6/23 & 135 & 6 & 3 & 50 & 0.37 \\
6/30 & 110 & 6 & 2 & 55 & 0.50 \\
7/7 & 87 & 7 & 4 & 52 & 0.60 \\
7/14 & 69 & 7 & 3 & 36 & 0.52 \\
7/21 & 102 & 2 & 1 & 9 & 0.09 \\
7/28 & 109 & 4 & 2 & 31 & 0.28 \\
8/4 & 70 & 4 & 2 & 36 & 0.51 \\
8/11 & 68 & 4 & 1 & 17 & 0.25 \\
8/18 & 63 & 2 & 1 & 28 & 0.44 \\
8/25 & 67 & 9 & 5 & 42 & 0.63 \\
9/1 & 78 & 3 & 2 & 45 & 0.58 \\
9/8 & 139 & 8 & 5 & 80 & 0.58 \\
9/15 & 124 & 9 & 9 & 91 & 0.73 \\
9/22 & 120 & 7 & 3 & 37 & 0.31 \\
9/29 & 84 & 2 & 1 & 27 & 0.32 \\
\hline
\end{tabular}
\caption{Weekly news clustering statistics (2024/5/26 - 2024/9/29) showing total news count, number of clusters, good (high-cohesion) clusters, clustered news (news in high-cohesion clusters), and the ratio of clustered to total news. }
\end{table}

\begin{figure}[htb]
\vspace{-5.9cm}
    \centering
    \includegraphics[width=0.47\textwidth]{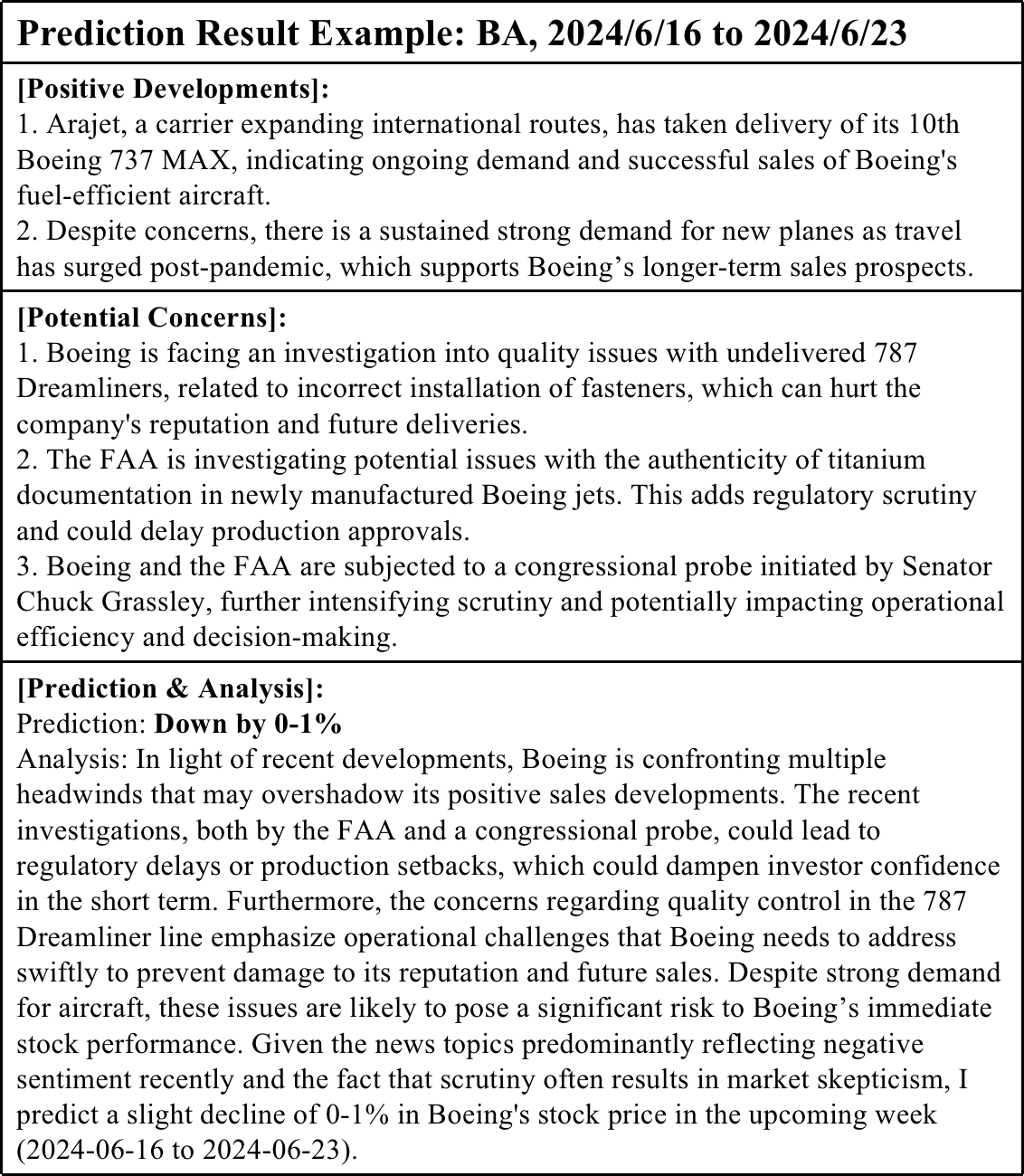}
    \caption{The prediction outcome of HG-NC method: Boeing Company, 2024/6/16-2024/6/23.}
    \label{prediction}
\end{figure}

\end{document}